\documentclass{article}
\usepackage{dingbat}
\usepackage{spconf,amsmath,graphicx}
\usepackage{amsfonts}
\usepackage{xcolor}
\usepackage{hyperref}
\usepackage{xspace}\usepackage[compact]{titlesec}         

\title{PIP: Physical Interaction Prediction via Mental Simulation with Span Selection}
\name{Jiafei Duan$^{1\ast}$ \qquad Samson Yu$^{2\ast}$ \qquad Soujanya Poria$^{2}$ \qquad Bihan Wen$^{3}$ \qquad Cheston Tan$^{1}$}

\address{\{duan\_jiafei, cheston-tan\}@i2r.a-star.edu.sg, \{samson\_yu, aspori\}@sutd.edu.sg, bihan.wen@ntu.edu.sg\\
$^{1}$Institute for Infocomm Research, A*STAR\\
$^{2}$Singapore University of Technology and Design\\
$^{3}$Nanyang Technological University of Singapore \\
}

\begin{document}
\setlength\abovedisplayskip{0pt}
\setlength\belowdisplayskip{0pt}
%
\maketitle
\begin{abstract}
Accurate prediction of physical interaction outcomes is a crucial component of human intelligence and is important for safe and efficient deployments of robots in the real world. While there are existing vision-based intuitive physics models that learn to predict physical interaction outcomes, they mostly focus on generating short sequences of future frames based on physical properties (e.g. mass, friction and velocity) extracted from visual inputs or a latent space. However, there is a lack of intuitive physics models that are tested on long physical interaction sequences with multiple interactions among different objects. We hypothesize that selective temporal attention during approximate mental simulations helps humans in physical interaction outcome prediction. With these motivations, we propose a novel scheme: \textbf{P}hysical \textbf{I}nteraction \textbf{P}rediction via Mental Simulation with Span Selection (PIP). It utilizes a deep generative model to model approximate mental simulations by generating future frames of physical interactions before employing selective temporal attention in the form of span selection for predicting physical interaction outcomes. To evaluate our model, we further propose the large-scale SPACE+ dataset of synthetic videos with long sequences of three prime physical interactions in a 3D environment. Our experiments show that PIP outperforms human, baseline, and related intuitive physics models that utilize mental simulation. Furthermore, PIP's span selection module effectively identifies the frames indicating key physical interactions among objects, allowing for added interpretability.
\end{abstract}

\def\thefootnote{*}\footnotetext{These authors contributed equally to this work}

\section{Introduction}
\label{sec:intro}

The ability to predict the outcomes of physical interactions among objects is a vital part of human intelligence \cite{moore2008mental, kubricht2017intuitive}. Yet, it is very challenging for AI systems to acquire this ability. The key to tackling this challenge lies in understanding commonplace physical events. AI systems need to possess this ability before they can be safely and efficiently deployed in the physical world \cite{duchaine2009safe,zheng2015scene,li2017visual}.

With the rapid advancements in computer vision, deep learning and embodied AI \cite{forsyth2011computer,bengio2021deep,duan2021survey}, there is an increase in intuitive physics models that aim to predict physical interaction outcomes. Many of these intuitive physics models \cite{battaglia2016interaction,wu2016physics,lerer2016learning,ye2018interpretable, groth2018shapestacks,leguen20phydnet} are inspired by the process of mental simulation from the \emph{mental physics engine} hypothesis in cognitive science research \cite{battaglia2013simulation,fischer2016functional,ullman2017mind,kubricht2017intuitive}. This hypothesis postulates that humans predict physical interactions via the process of mental simulation. With only a few initial visual inputs of physical interaction, we can mentally reconstruct the scene with some initial approximations of the physical proprieties and dynamics of the objects. We can then predict the outcomes of physical interactions using this estimated information and the generated future visual states of objects during mental simulation. However, existing intuitive physics models are tested on short video sequences from datasets with mostly one continuous physical interaction among objects. Furthermore, it is uncertain whether humans can estimate physical properties accurately from visual inputs, and whether accurate physical property prediction is always useful for predicting physical interaction outcomes. In some cases, despite biases in estimations of physical properties \cite{fleming2014visual,rossi2018speed,mitko2021striking}, humans have been found to have adequately precise physical interaction outcome predictions \cite{mitko2020all}. This suggests that humans might have other cognitive abilities on top of
the physical property estimation that enable good physical interaction outcome prediction.

Past research has shown that humans make rational probabilistic inferences about physical interaction outcomes in a ``noisy Newtonian" framework, assuming Newton's laws plus noisy observations \cite{battaglia2013simulation}. We use noisy and approximate physical simulations to account for property, perceptual, dynamic and even collision uncertainties \cite{battaglia2013simulation, gerstenberg2017intuitive, Hamrick2015ThinkAT, Bramley2018IntuitiveEI, LUDWINPEERY2021101396, smith2013sources, ullman2017mind}. We posit that one of the beneficial cognitive abilities in humans for effective physical interaction outcome prediction is the ability to perform mental simulation with selective temporal attention to focus on physically relevant moments \cite{firestone2016seeing}. This might be because noisy observations and simulations are counterproductive except in moments when crucial physical interactions (e.g. collision events \cite{LUDWINPEERY2021101396, ullman2017mind, gerstenberg2017intuitive}) are present. We further posit that the selected moments in the mental simulation are then used to finalize outcome predictions.

Inspired by our hypothesis that humans use selective temporal attention in noisy mental simulations to reduce the negative effects of noise, we propose PIP, an intuitive physics model with future frame generation and span selection for predicting physical interaction outcomes. The span selection module serves as the temporal attention mechanism to focus on key physical interaction moments in the generated frames. Since state-of-the-art generative models in video generation still have artifacts and prediction errors in their generations \cite{weissenborn2019scaling, yan2021videogpt}, we simply use the well-established convolutional LSTM (ConvLSTM) \cite{xingjian2015convolutional, guen2020disentangling} for future frame generation to approximate noisy mental simulations as a starting point. 

Our contributions include: (a) PIP, a novel intuitive physics model for effective predictions of physical interaction outcomes among objects in long sequences with disjointed object interactions, and (b) the SPACE+ dataset, the first synthetic video dataset with long sequences of multiple disjointed object interactions for three fundamental physical interactions (\emph{stability, contact} and \emph{containment}) in a 3D environment. (c) Our experiments show that PIP outperforms existing intuitive physics models and humans in predicting the outcomes of physical interactions while identifying key physical interaction moments.

\begin{figure*}[h]
    \centering
    \includegraphics[width=\textwidth]{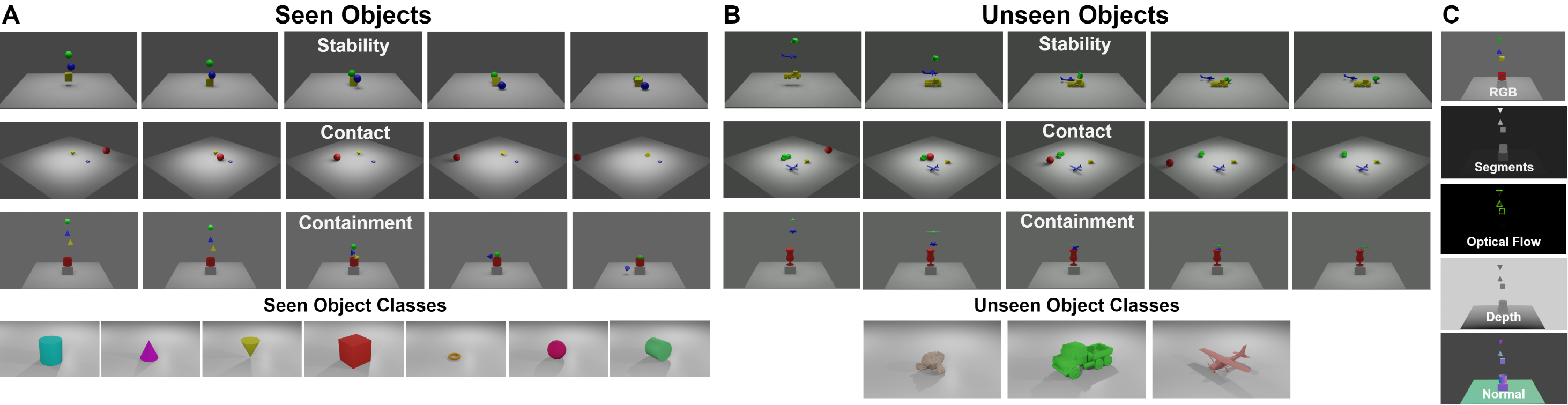}
    
    \caption{Examples from SPACE+ dataset: (A) Frames of the three physical interaction tasks from the SPACE+ dataset for the seen object scenario. (B) Frames of the same tasks with new object classes for the unseen object scenario. (C) Visual information for one frame: RGB, object segmentation, optical flow, depth and surface normal vector.}
\label{fig:1}
\end{figure*}

\begin{table}
\begin{center}
    \begin{tabular}{l|c|c}
    \hline
    Physical Interactions & Scenarios & Frames \\
    \hline\hline
    Stability & 19,551 & 2,932,650\\
    Contact & 19,551 & 2,932,650\\
    Containment & 17,955 & 2,693,250 \\
    \hline
    Total & 57,057 & 8,558,550\\
    \hline\hline
    Unseen Stability & 3910 & 586,500\\
    Unseen Contact & 3910 & 586,500\\
    Unseen Containment & 3591 & 538,650 \\
    \hline
    Total & 11,411 & 1,711,650\\
    \hline
    \end{tabular}
\end{center}
\caption{SPACE+ dataset analysis.}
\label{dataanalysis}
\end{table}

\section{Related Work}
\label{sec:Related Work}
Several synthetic video datasets based on fundamental physical interactions among objects in 3D environments have been developed \cite{ye2018interpretable,groth2018shapestacks,bear2021physion,duan2021space,dasgupta2021avoe} with the growing importance of 3D simulation work in AI \cite{duan2021survey}. As a result, a diverse range of intuitive physics models \cite{brubaker2009estimating,fragkiadaki2015learning,lerer2016learning,li2016fall,finn2016unsupervised,duan2021space} were created for predicting outcomes of physical interactions among objects. However, we find Physics 101 \cite{wu2016physics}, Interpretable Intuitive Physics Model \cite{ye2018interpretable}, and PhyDNet \cite{baradel2019cophy} to be the most relevant to our work as they include mental simulation in their intuitive physics models, and their models are trained on video datasets with physical interactions.

\textbf{Physics 101} \cite{wu2016physics} introduced a video dataset containing over 101 real-world physical interactions of objects in four different physical scenarios. It further proposed an unsupervised representation learning model to tackle the Physics 101 dataset. The model learns directly from unlabeled videos to output the estimates of physical properties of objects, and the generative component of the model can then be used for predicting the outcomes of physical interactions.

\textbf{Interpretable Intuitive Physics Model} \cite{ye2018interpretable} proposed an encoder-decoder framework for predicting future frames of collision events. The encoder layers will extrapolate the physical properties such as mass and friction from the input frames. The decoder then disentangles latent physics vectors by outputting optical flow. For a collision event, a bilinear grid sampling layer takes the optical flow and the input frames to produce a prediction of its outcome in the form of a future frame. The dataset used for training the model is a synthetic video dataset of collision events with 11 different object combinations of 5 unique basic objects generated using the Unreal Engine 4 (UE4) game engine.

\textbf{PhyDNet \cite{guen2020disentangling}} leverages the physical knowledge extracted from partial differential equations (PDE) to improve unsupervised video prediction on videos with physical interactions and dynamics. PhyDNet does so in a two-branch approach. PhyDNet's architecture separates the PDE dynamics from unknown complementary information. PhyCell, a deep recurrent physical model, performs PDE-constrained predictions for PDE dynamics, while a ConvLSTM \cite{xingjian2015convolutional} is used to model the complementary information. PhyDNet outperforms state-of-art methods in unsupervised video prediction of physical interaction outcomes.

The related works focus primarily on extracting physical properties of objects and dynamics from visual inputs for generating future frames, which are later used for predicting the outcome of physical interactions. In this work, we propose a new direction for mental simulation in predicting physical interaction outcomes by incorporating selective temporal attention. Our method first generates the future frames to model approximate mental simulation, then uses span selection to focus on key moments in the simulation. 

\section{SPACE+ Dataset}
\label{sec:dataset}

The proposed SPACE+ dataset, an improved extension of the SPACE dataset \cite{duan2021space}. The original SPACE dataset comprises three novel video datasets synthesized by the SPACE simulator from 3D scenarios based on three fundamental physical interactions: stability, contact and containment. The SPACE dataset allows for the configuration of several parameters such as object shapes, the number of objects, object spawn locations and container types (only applicable to the containment task) during the generation.

The SPACE dataset has 15,000 unique scenarios with 5,000 scenarios for each of the three tasks. From there, 15,000 videos are generated lasting 3 seconds each at a frame rate of 50 frames per second (FPS), adding up to 2 million frames. The SPACE dataset is balanced in the outcome of its physical interactions on individual objects except for the stability task, which is inherently unbalanced. As during stability, for scenarios where two or three objects are spawned and land on top of each other, the objects that spawn above other will have higher chances of being unstable.

Our SPACE+ dataset expands and improves upon the existing SPACE dataset. Without altering the adjustable parameters, we further generate 42,057 unique scenarios on top of the original 15,000 scenarios created using the SPACE simulator as shown in Table \ref{dataanalysis}. These scenarios follows the data distribution of the original SPACE dataset. We collect up to 57,057 videos with over 8 million frames in total. Beyond scaling up the size of the dataset, we also add new object classes for all three fundamental physical interactions in the SPACE+ dataset, as shown in Figure \ref{fig:1}B. These new object classes will be used in an unseen object scenario that will help us evaluate the generalizability of our models and human performance, since unexpected physical interactions might arise due to the new and complex shapes of the new objects. The original object classes $O=$ \{\emph{cylinder, cone, inverted cone, cube, torus, sphere, flipped cylinder}\} in the SPACE dataset are shown in Figure \ref{fig:1}A, and will be used in the seen object scenario, i.e. our models and humans will be able to train on or familiarize themselves with these object classes in the various physical interaction tasks before predicting their physical interaction outcomes in the tasks. For the SPACE+ dataset, besides the RGB frames, we also follow the SPACE paper in providing the object mask, segmentation map, optical flow map, depth map and surface normal vector map, as shown in Figure \ref{fig:1}C.

\begin{figure*}[h]
    \centering
    \includegraphics[width=\textwidth]{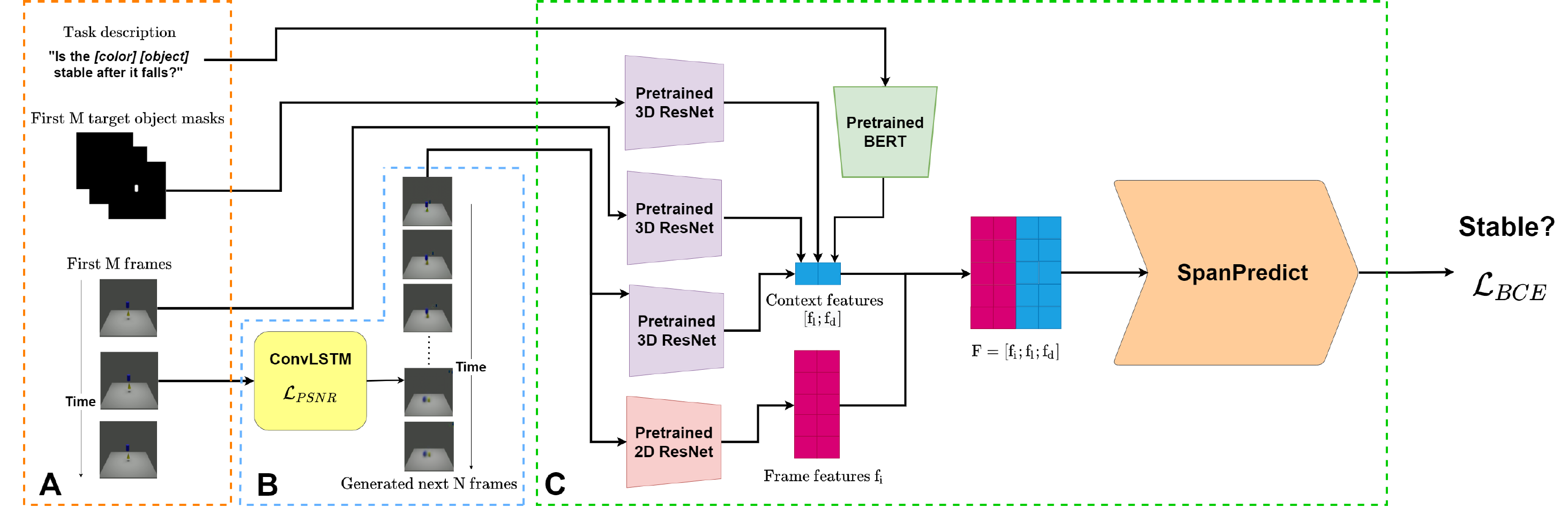}
    \caption{PIP model architecture. (A) \textbf{Data inputs}: the original data inputs for our physical interaction prediction task comprise of the first M frames, the first M target object masks and the task description. The task description is different for each of the three fundamental tasks to facilitate multi-task learning for the combined task (in this diagram the task description for the stability task is shown). (B) \textbf{Mental simulation}: the first M frames are fed into the mental simulation module that consists of a ConvLSTM to generate the next N frames. (C) \textbf{Span selection}: the original data inputs and the generated N frames are fed into the span selection module, where pretrained models will encode them into features before classification. All models are trained. }
    \label{fig:architecture}
\end{figure*}

\section{PIP}
\label{sec:pip}

As shown in Figure \ref{fig:architecture}, PIP utilizes a ConvLSTM for future frame prediction to mimic noisy mental simulations and span selection to incorporate selective temporal attention. 2D/3D residual networks (ResNets) \cite{He_2016_CVPR, hara2018can, kataoka2020would} and a pretrained BERT \cite{devlin2018bert} are used to encode the necessary visual and task information for span selection \cite{subramanian2021spanpredict}. PIP enables interpretability through span selection without the costly and subjective frame-based annotations needed for typical key frame selection approaches \cite{s20236941}.

\subsection{Mental Simulation}
We use a ConvLSTM for future frame prediction to mimic noisy mental simulation as it is well-established and forms the backbone of recent video prediction approaches \cite{zhang2019spatio, chai2021cms}. The input frames are individually encoded into features with convolutional and transposed convolutional layers modelled after deep convolutional generative adversarial networks (DCGAN) \cite{radford2015unsupervised, guen2020disentangling} and individually fed as inputs into the ConvLSTM in sequence. We make use of teacher forcing \cite{6795228} where we provide ground-truth frames to the model instead of generated frames to improve model learning. Starting from a specified frame, we train the ConvLSTM for future frame prediction.

A peak signal-to-noise ratio (PSNR) loss is used to train the convolutional layers and the ConvLSTM. The weights from each of them are shared across all three tasks in the combined training scenario.

\subsection{Span Selection}
PIP includes a span selection module to focus on salient frames while learning to predict physical interaction outcomes. Span selection allows for added interpretability by identifying the frames that are important for physical interaction prediction. Furthermore, this is done without labor-intensive frame-based annotations.

We use SpanPredict, a model proposed by \cite{subramanian2021spanpredict} in natural language processing (NLP) for document classification with only classification labels in the absence of ground-truth spans. In this work, we focus on physical interaction outcome prediction in videos with only classification labels in the absence of ground-truth spans.

We modify SpanPredict to take in features for each generated frame. We obtain image features for each frame $\mathrm{\textbf{f}}_{i,t}\in\mathbb{R}^{i}$ by passing them down a pretrained 2D ResNet50.

To facilitate multi-task learning in the combined task and standardize inputs for all four tasks, we encode different language features for each of the three fundamental tasks. This helps to prevent model size from increasing with the number of tasks. For the stability, contact and containment tasks, we create the queries \emph{``Does the [color] [object] get contacted by the red ball?}", \emph{``Is the [color] [object] contained within the containment holder?}" and \emph{``Is the [color] [object] stable after it falls?}" respectively for each object in a scene. We obtain subword tokens for a query $Q=\{q_1, q_2,...,q_n\}$ using the WordPiece tokenizer \cite{wu2016google} and process the sequence as $[CLS]\ Q\ [SEP]$, as per the standard format for single sentence inputs into BERT models. We then feed the processed sequence into a pretrained BERT model (specifically bert-base-uncased). The language features $\mathrm{\textbf{f}}_l\in\mathbb{R}^{l}$ are derived from the embeddings corresponding to the $[CLS]$ token, and are concatenated to each generated frame's feature.

In addition, for each generated frame's feature, we concatenate features $\mathrm{\textbf{f}}_d\in\mathbb{R}^{d\times3}$ from the first 3 frames, the first 3 segmentation masks for the target object and all generated frames. We pass each of them through a different pretrained 3D ResNet34 to get their features. Intuitively, these 3 sources of information provide the model with prior knowledge, object tracking and global contextual information respectively. The combined features for each generated frame $\mathrm{\textbf{f}}_t = [\mathrm{\textbf{f}}_{i,t};\mathrm{\textbf{f}}_{l};\mathrm{\textbf{f}}_{d}]$ are stacked to form a sequence of features $\mathrm{\textbf{F}}=[\mathrm{\textbf{f}}_1,\mathrm{\textbf{f}}_2,...,\mathrm{\textbf{f}}_T]$.

In the following paragraph, we will briefly explain SpanPredict. We set the number of spans, and for each span we provide a pair of trainable attention weights, $\mathrm{\textbf{w}}_p,\mathrm{\textbf{w}}_q\in\mathbb{R}^{i+l+d\times3}$. This allows for flexibility for physical interactions that might require two or more disjointed segments in a video to determine their occurrence. Using these attention weights, we get vectors $\tilde{\mathrm{\textbf{p}}} = \mathrm{softmax(\textbf{F}^T \textbf{w}_p)}$ and $\tilde{\mathrm{\textbf{q}}} = \mathrm{softmax(\textbf{F}^T \textbf{w}_q)}$, which represent the set of probabilities of each frame being the start and end of a salient span respectively. We then produce a span representation $\mathrm{\textbf{r}}$ for each span using the cumulative sum function. Specifically, we sum up the set of probabilities for each span cumulatively, i.e. cumulative sum, such that $\mathrm{\textbf{p}}=\mathrm{cumsum}(\tilde{\mathrm{\textbf{p}}})$ and $\mathrm{\textbf{q}}=\mathrm{cumsum}(\tilde{\mathrm{\textbf{q}}}_{::-1})$, where $\tilde{\mathrm{\textbf{q}}}_{::-1}$ is $\tilde{\mathrm{\textbf{q}}}$ with its elements reversed. Intuitively, each element in $\mathrm{\textbf{p}}$ and $\mathrm{\textbf{q}}$ represents the probability that the start of a span has occurred by that element when coming from the left of the sequence and the probability that the end of a span has occurred by that element when coming from the right of the sequence respectively. We then combine both start and end positional information as $\tilde{\mathrm{\textbf{r}}} = \mathrm{\textbf{p}}\odot \mathrm{\textbf{q}}$ to assign larger weights to frames that have high mass under both $\mathrm{\textbf{p}}$ and $\mathrm{\textbf{q}}$, i.e. those that fall between the start and end points of a span. Finally, we normalize $\tilde{\mathrm{\textbf{r}}}$ such that its elements sum to 1:
\begin{displaymath}
\mathrm{\textbf{r}}=\frac{\mathrm{\textbf{p}}\odot \mathrm{\textbf{q}}}{\Sigma_t(\mathrm{\textbf{p}}\odot \mathrm{\textbf{q}})_t + \epsilon},
\end{displaymath}
where $\epsilon$ is a small constant. $\mathrm{\textbf{r}}$ gives us the final score of each frame's contribution to the span. We weigh the combined features $\mathrm{\textbf{F}}$ by $\mathrm{\textbf{r}}$, then average its values across its temporal dimension to get $\mathrm{\textbf{m}}=\mathrm{average}(\mathrm{\textbf{Fr}}) \in \mathbb{R}^{i+l+d\times3}$. To get the overall span score for a span, i.e. its contribution to the final classification, we use a third attention weight $\mathrm{\textbf{w}}_z$ to get $\mathrm{\textbf{z}}=\mathrm{\textbf{mw}}_z$, and we repeat this process for every span with the same $\mathrm{\textbf{w}}_z$. Finally, the span scores for all spans are summed up and passed through a sigmoid layer to predict $\hat{y}\in\{0,1\}$. An additional explicit penalty is also included in the form of the generalized Jensen-Shannon divergence \cite{61115} to make the spans more concise and distinct (i.e. minimize overlapping frames for multiple span selections) \cite{subramanian2021spanpredict}.

\begin{figure*}[t]
\centering
\includegraphics[width=\linewidth]{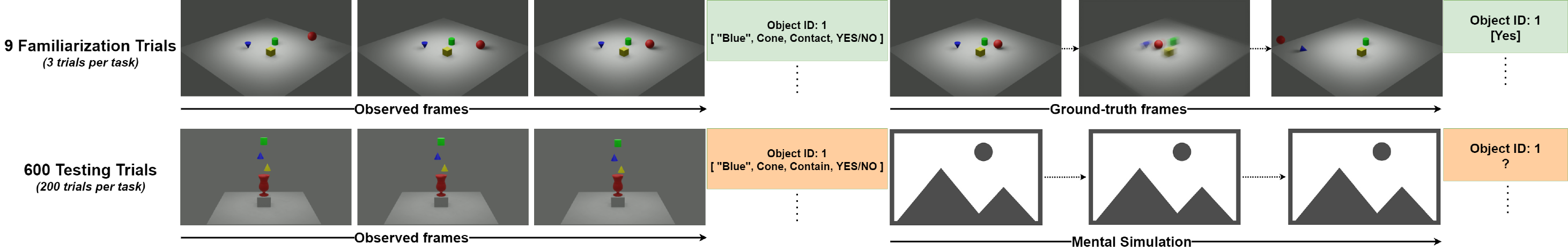} 
\caption{Human trial setup on physical interaction prediction tasks. Trial structure for familiarization trials (\emph{top}) and test trials (\emph{bottom}) with the observed frames, task queries and ground-truth frames.}
\label{human}
\end{figure*}

\begin{table*}[t]
\begin{center}
    \begin{tabular}{l|cccc|cccc}
    \hline
     & \multicolumn{4}{c}{Seen Objects (\%)} & \multicolumn{4}{c}{Unseen Objects (\%)}\\
    \hline
    Methods & {Stability} & {Contact} & {Containment}  & {Combined}  & {Stability} & {Contact} & {Containment}  & {Combined}\\
    \hline\hline
    Human  & 80.38 & 61.80 & 78.30 & 72.00 & 64.03 & 55.94 & \textbf{77.08} & 62.08\\
    Baseline & 92.35 & 65.63 &  78.70 & 60.78& 63.17 & 54.06 &  55.47 & 54.98 \\
    PhyDNet  & \textbf{92.36} & 61.23 &  79.47 & 62.03& 63.78 & 55.41 &  59.10 & 55.61 \\
    PIP w/o SS  & 92.18 & 61.84 &  80.39 & 58.70& \textbf{67.07} & 52.14 & 60.27 & 54.59 \\
    PIP (Ours) &92.33 & \textbf{87.50} & \textbf{86.45} & \textbf{77.71} & 66.41 & \textbf{56.33} & 55.99 &  \textbf{62.23}\\
    \hline
    
    \end{tabular}
\end{center}
\caption{Accuracy results for seen (\emph{left}) and unseen (\emph{right}) object scenarios for all four physical interaction outcome prediction tasks.}
\label{ta1}
\end{table*}

\section{Experiments}
\subsection{Experimental Setup}
The SPACE+ dataset is divided into stability, contact and containment tasks, and we further create a new combined task that contains an equal number of samples from each of the three fundamental tasks.

For each of the three fundamental tasks and the combined task, we use 1,000 scenes from the SPACE+ dataset that is representative of the full dataset and split it into 60\% for the training set, and 20\% for the validation and test set each. For the combined task, each of the splits has equal numbers of each of the three fundamental tasks to ensure that the dataset is balanced. For each scene, the physical interaction prediction is done for individual objects, i.e. the inputs are the same across objects in the same scene except for the object masks, and the labels are different across objects. To compare the models' performance with human performance using the same number of samples in the test set, we do not use the full SPACE+ dataset for training to maintain the training-validation-test split ratios. 

For the unseen object scenario introduced with SPACE+, we also take 200 scenes for each of the three fundamental tasks and the combined task. The combined task contains equal numbers of each of the three fundamental tasks.

For each scene, there is a 3-second video with a FPS of 50 to make up 150 total frames. To limit the size of the dataset to improve computational runtime, we use a frame interval of 2 where we skip 1 frame every 2 frames, resulting in 75 frames in total. Of these 75 frames, we take the first 3 frames as initial frames to be shown to both human subjects and PIP, since there are no physical interactions among objects in these first 3 frames (i.e. first 6 frames in the original frame sequence with a frame interval of 1) in all scenes. For the ConvLSTM, we provide 37 subsequent frames from the 75 frames in addition to 3 initial frames to train it to learn future frame prediction. This is because 40 frames with a frame interval of 2 (i.e. 80 frames in the original frame sequence with a frame interval of 1) allow all outcomes of physical interactions among objects to be known, while having fewer frames reduces computational runtime. For each object in a scene, there are also 150 segmentation masks indicating its location in the 150 frames.

\subsection{Evaluation Metric}
To evaluate the performance of our selected methods, we use classification accuracy of physical interaction outcome predictions on the test sets of both seen and unseen object scenarios:
\begin{displaymath}
\mathrm{score} = \begin{cases}
1\quad \mathrm{if}\ \hat{y} = y\\
0\quad \mathrm{otherwise},
\end{cases}
\end{displaymath}
where $y\in\{0,1\}$ is the ground-truth label.

    

\begin{figure*}[t]
\centering
\includegraphics[width=\linewidth]{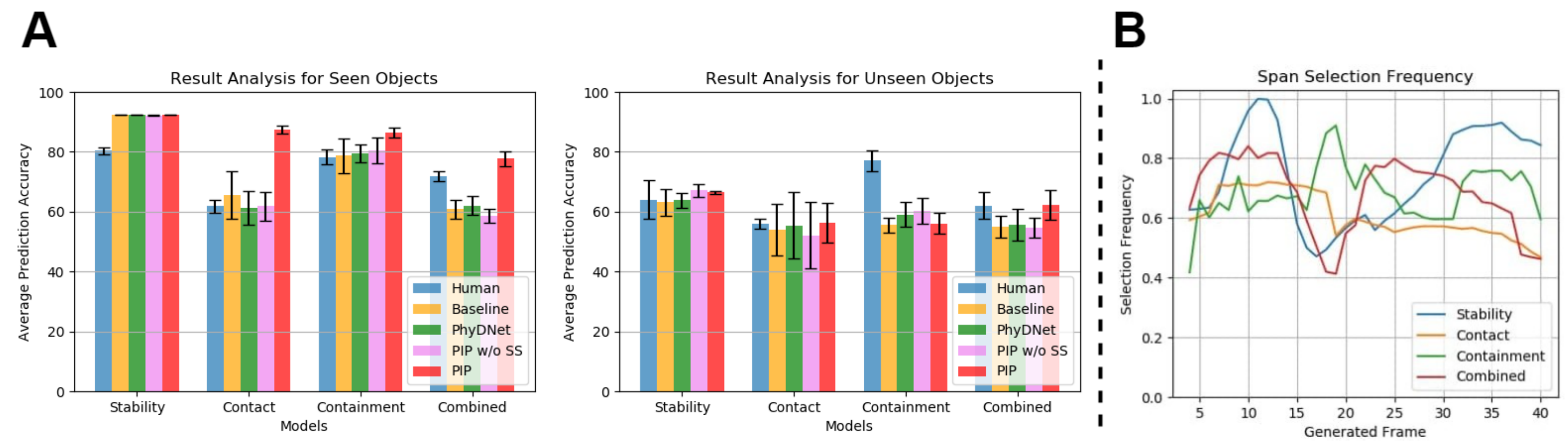} 
\caption{(A) Average test prediction accuracy and standard deviation for seen (\emph{left}) and unseen object (\emph{right}) scenarios across all models and seeds. (B) PIP's frame selection frequencies on the test set for seen object scenarios across all seed runs.}
\label{humanresult}
\end{figure*}

\begin{figure*}[t]
\centering
\includegraphics[width=\linewidth]{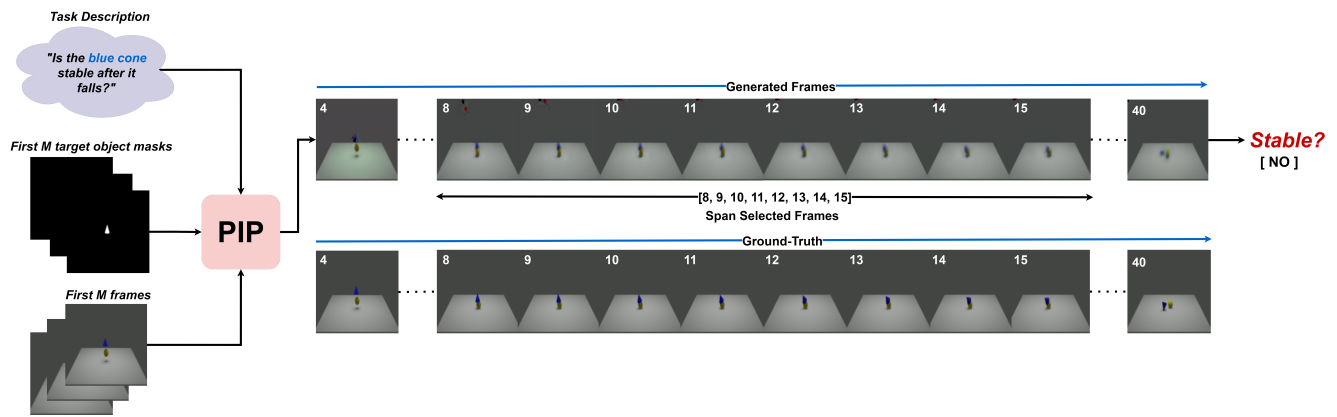} 
\caption{An example of PIP's generation and span selection corresponding to the first window of peak span selection frequencies in the stability task. For visualizations of key physical interaction moments in the other tasks, refer to our supplementary material.}
\label{span}
\end{figure*}



\subsection{Human Baseline}
We conduct a simple human experiment to obtain a benchmark for human performance in predicting physical interaction outcomes. Five participants were recruited anonymously from the internet for the experiments. The participants first undergo a familiarization trial with nine questions (three questions for each of the three physical interaction tasks) for only the seen objects. In each familiarization trial, the participants are first shown a video with three continuous observed frames containing the initial moments of a physical interaction scenario.  The participants are then asked to predict the outcome of the physical interaction by indicating either "YES" or "NO" for specified objects as shown in Figure \ref{human}. After the participants have completed indicating their prediction, they are shown the remaining parts of the video and are thus able to evaluate their predictions. After completing the familiarization trials, they proceed to the actual test trials beginning with the scenarios with seen objects and then the scenarios with unseen objects. The test trials are similar to the familiarization trial, but the full videos are not revealed to the the participants at the end of each submission. Upon completion, their results are computed and they are informed of their task-specific accuracy and standard deviation for both the seen and unseen object scenarios.


\subsection{Baseline}
We establish baseline performance by building a model similar to PIP without the mental simulation and span selection modules. This baseline model takes in the first 3 frames, the first 3 segmentation masks and the BERT language features, and encodes the frames and the segmentation masks with separate pretrained 3D ResNet34s. It then uses linear layers for classification of the concatenated features. The results, averaged across 5 runs with different seeds, are shown in Table \ref{ta1} as ``Baseline".

\subsection{PhyDNet}
We modify PhyDNet \cite{guen2020disentangling} for a performance comparison between our approach and the incorporation of physical dynamics in mental simulations. Like PIP, this modified PhyDNet model takes in the first 3 frames, the first 3 segmentation masks and the BERT language features, and encodes the frames and the segmentation masks with separate pretrained 3D ResNet34s. However, this model generates 37 subsequent frames using the first 3 frames with the PhyDNet model instead of a ConvLSTM. It then uses another pretrained 3D ResNet34 to encode the 37 frames as features, before all the features are concatenated, and linear layers are used for classification of the combined features. The results, averaged across 5 runs with different seeds, are shown in Table \ref{ta1} as ``PhyDNet".

\subsection{PIP}
\subsubsection{Implementation Details}
Models are implemented using PyTorch \cite{NEURIPS2019_9015} and BERT is implemented using Hugging Face's Transformers \cite{wolf2019huggingface}. We train using the Adam optimizer \cite{kingma2014adam} for 20 epochs with a fixed learning rate of 1e-3 and a batch size of 2. We set a teacher forcing \cite{6795228} rate of 0.1 for the ConvLSTM. Our ConvLSTM module consists of 3 ConvLSTM layers, 6 convolutional layers and 6 transposed convolution layers. We fine-tune the pretrained 2D ResNet50, 3D ResNet34s and BERT. We use PSNR loss to train the ConvLSTM and binary cross-entropy loss to train the entire model. To decide when to stop training, we monitor the validation classification accuracy for physical interaction outcome prediction. Our best model is selected based on its classification accuracy on the validation set of the seen object scenario. We train and test PIP over 5 runs with different random seeds and average the results. The number of spans extracted is set to 3. Experiments were run across NVIDIA GPUs (RTX A6000, V100 and GeForce RTX 2080 Ti) on Linux servers and the total time for each epoch is about 2-3 hours for a total of about 40-60 hours for the entire process of 20 epochs.

\subsubsection{Ablation Study}
We conduct an ablation study to examine the effect of the span selection module. Like PIP, this ablation model takes in the first 3 frames, the first 3 segmentation masks and the BERT language features, and encodes the frames and the segmentation masks with separate pretrained 3D ResNet34s. Using the first 3 frames, this ablation model also generates 37 subsequent frames with a ConvLSTM. However, it uses another pretrained 3D ResNet34 to encode the 37 frames as features, before all the features are concatenated, and linear layers are used for classification of the combined features. We use the same hyperparameters and seed runs to train and test this model as those for PIP. The results, averaged across 5 runs with different see, are shown in Table \ref{ta1} as ``PIP w/o SS".

\section{Results and Analysis}

\subsection{Human Performance}
Based on the results obtained from the human experiments, as shown in Table \ref{ta1} and Figure \ref{humanresult}A, we observe that human performance is consistent with an average standard deviation of 1.87 across all tasks in the seen object scenario. Human performance is also affected by the complexity of the shapes of the objects. Human performance has an average decrease of 8.34\% from the seen object scenario to unseen object scenario across all tasks and a higher average standard deviation of 3.96, suggesting lower consistency. However, it has the lowest decrease when compared to the other models, suggesting the generalizability of human performance. Furthermore, human performance for the containment task in the unseen object scenario has the lowest decrease of 1.22\% and is significantly higher than that of the other model methods by a difference of at least 16.81\%. We believe this anomaly is due to the fact that humans can employ heuristics based on the estimation of physical properties (e.g. the width of the unseen object in comparison to the width of the container's entrance determines containment success) to improve predictions in the containment task. This ability is a known complement to the human mental simulation process \cite{ullman2017mind}.

\subsection{PIP Test Performance}

Based on our test results from Table \ref{ta1}, PIP achieve accuracies of 92.33\% for \emph{stability}, 87.50\% for \emph{contact}, 86.45\% for \emph{containment} and 77.71\% for \emph{combined} in the seen object scenario. PIP surpasses human performance by an average of 12.87\% across all tasks with seen objects, the baseline model by 11.63\% and the ablation model by 12.71\%. PIP also surpasses the modified PhyDNet by 16.31\% excluding for the stability task, where PIP performs slightly worse. Furthermore, the average standard deviation of PIP is only 1.36, which is the lowest compared to other models and human performance. Lastly, PIP is the only model that outperforms human performance in all four tasks in the seen object scenario and three tasks in the unseen object scenario. 

The results suggests that PIP is effective in predicting the outcomes of physical interactions, as it outperforms most of the models significantly in seen object scenarios with a double-digit margin for some tasks. PIP is also highly consistent in its prediction accuracy for all tasks. Moreover, through comparison of PIP with the ablation results, it can be seen that the span selection mechanism improves the prediction performance. The only anomaly is in stability performance. For the stability task, the results between the different models are relatively close. These high accuracy predictions for the stability task could indicate ``shortcut learning" \cite{geirhos2020shortcut} rather than the model learning the physical understanding behind object interactions. Furthermore, the performance for the stability task is also significantly higher than the other tasks. We hypothesize that the high performance of stability is partly due to the imbalance of physical states as mentioned in Section \ref{sec:dataset}. However, this imbalance in the distribution of object physical states in the stability tasks reflects an accurate representation of real-world physical dynamics where there is a higher probability of instability in multiple object scenarios. This is further supported by our analysis of the SPACE+ dataset in the supplementary material. We also evaluate the generalizability of the models by testing them on unseen object scenarios. We show in Table \ref{ta1} that all the methods perform more closely in relation to one another in unseen object scenarios with an average difference of not more than $\pm$7.25\% except for human performance in the containment task.


\subsection{Span Selection}

Based on our results, we show that PIP outperforms our ablation model by a huge average margin of 12.71\%. This supports our premise that span selection as a form of selective temporal attention helps mental simulations to improve predictions of physical interaction outcomes. PIP also outperforms the modified PhyDNet model by 16.31\%, suggesting that PIP contributes more to mental simulations than the incorporation of learned physical dynamics. Furthermore, PIP's span selection allows us to understand how important each frame is in its contribution to physical interaction outcome predictions, providing added interpretability. This added interpretability is a significant advantage as it gives us greater insights on how to improve model performance and build trust in applications where safety is a priority.

In our experiments, it is difficult for the model to follow a strict threshold of 0 for $\mathrm{\textbf{r}}$ during salient frame selection. It is also difficult to use a static threshold, even if it is normalized by generated frame sequence length $N$ (i.e. $\frac{1}{N}$), due to the way $\mathrm{\textbf{r}}$ is calculated. Hence, we propose a new way to calculate the threshold for $\mathrm{\textbf{r}}$ as such:
\begin{gather*}
\mathrm{\textbf{p\_{threshold}}} = \mathrm{cumsum}([\frac{1}{N},\frac{1}{N},...,\frac{1}{N}]) \in \mathbb{R}^N \\
\mathrm{\textbf{q\_threshold}} = \mathrm{\textbf{p\_threshold}_{::-1}} \\
\mathrm{\textbf{r\_threshold}} = \frac{\mathrm{\textbf{p\_{threshold}}} \odot \mathrm{\textbf{q\_{threshold}}}}{\Sigma_t(\mathrm{\textbf{p\_{threshold}}}\odot \mathrm{\textbf{q\_{threshold}})_t}},
\end{gather*}
where $N$ is the generated frame sequence length. Intuitively, this sets a uniform distribution normalized by sequence length as the threshold for $\mathrm{\textbf{p}}$ and $\mathrm{\textbf{q}}$ and calculates the threshold for $\mathrm{\textbf{r}}$ in the same way  $\mathrm{\textbf{r}}$ is obtained from $\mathrm{\textbf{p}}$ and $\mathrm{\textbf{q}}$. We show in Figure \ref{span} an example of PIP's effective selection of salient frames using this threshold calculation.

In Figure \ref{humanresult}B, we show the frequencies of each generated frame being selected across all four tasks and all five seed runs, from frame 4 to frame 40, for the seen object scenario. Furthermore, upon inspection of the generated frames, we found that peaks in Figure \ref{humanresult}B indicate moments of key physical interactions among objects. For example, in Figure \ref{span}, we illustrate for the stability task that frames 8-13, which corresponds to the first peak in the stability task's span selection frequencies, capture the first physical interactions among the ground and the falling object(s).

The span selection frequencies also highlight the complexity of each task. For example, the stability task's selected frames are mainly distributed into two distinct windows of frames 8-13 and 30-40 with high frequencies. This suggests that the stability task has two consistent and distinct moments of key physical interactions, which might allow for overfitting from generative models if they focus on these moments. On the other hand, for the contact task with a balanced distribution of selected frames, overfitting is more difficult. This highlights PIP's significance, since it outperforms all other models significantly for the contact task in the seen object scenario.

Finally, for the combined task, the span selection frequencies generally follow the trends of the three fundamental tasks in the first part before frame 15 with a small peak. The frequencies after frame 15 generally follow those of the contact task. More importantly, at frames 18 and 19, the frequencies decrease significantly in stark contrast to the containment task. Furthermore, the frequencies show a decreasing trend after a peak at frame 25, in contrast to the stability and containment tasks. This suggests that the combined task helps PIP to learn novel features that allow for generalizability across the three fundamental tasks. These features improve PIP's robustness, as seen in Table \ref{ta1} where PIP has lowest accuracy decrease of 15.48\% from the seen object scenario to the unseen scenario for the combined task, whereas there is a decrease of at least 25.92\% for the other tasks.

\section{Limitations and Future Work}
PIP currently performs worse on each of the three fundamental tasks when it is trained on the combined task than when it is both trained and tested on each of the three tasks. This is a common problem in multi-task learning \cite{standley2020tasks}. Since PIP is only trained on a small number of samples for each of the three tasks in the combined task scenario, future work can be done with more data to improve PIP's robustness in multi-task settings. Furthermore, computation time increases with the number of input and generated frames. Lastly, we also acknowledge that more representative results for the human baseline performance could have been obtained if we had more participants for the human experiments.

We make two key assumptions. We assume that the differences in performance between the different models will remain the same even if we train on the full 5,000 scenes in the SPACE+ dataset instead of 1,000 scenes for each scenario. We also assume that the generation artifacts and errors in the ConvLSTM's generations accurately model ``noisy Newtonian" dynamics in human mental simulations. Future work can be done to better model ``noisy Newtonian" dynamics in our model's mental simulations and test how the precision in approximations of different physical properties (e.g. size and shape) in mental simulations affects model performance on different physical interaction tasks.

\section{Conclusion}
Our ability as humans to effectively predict the outcomes of physical interactions among objects in the real world is crucial to ensure safety and success in performing complex tasks. Similarly, it is important for AI systems to develop this capability. This intuitive understanding of commonplace physical interactions is critical for complex real-world tasks such as human-robot collaboration and self-driving cars, which require reacting to ever-changing physical dynamics. In this work, we propose a new direction for intuitive physics models by proposing PIP, an intuitive physics model with selective temporal attention via span selection to improve physical interaction outcome prediction in noisy mental simulations. We evaluate PIP on the SPACE+ dataset, and show that PIP outperforms baseline and related intuitive physics models, and human performance, while identifying key physical interaction moments and providing added interpretability with span selection.

\section*{Acknowledgments}
This research is supported by the Agency for Science, Technology and Research (A*STAR), Singapore under its AME
Programmatic Funding Scheme (Award \#A18A2b0046) and the National Research Foundation, Singapore under its NRF-ISF Joint Call (Award NRF2015-NRF-ISF001-2541). We would also like to thank the Machine Learning and Data Analytics Lab at School of Electrical and Electronic Engineering, Nanyang Technological University of Singapore, Rapid-Rich Object Search (ROSE) Lab, Nanyang Technological University of Singapore and DeCLaRe Lab, Singapore University of Technology and Design for the computational resources used for this work.

\bibliographystyle{IEEEbib}
\bibliography{strings,refs}

\begin{thebibliography}{10}

\bibitem{moore2008mental}
David~S Moore and Scott~P Johnson,
\newblock ``Mental rotation in human infants: A sex difference,''
\newblock {\em Psychological science}, vol. 19, no. 11, pp. 1063--1066, 2008.

\bibitem{kubricht2017intuitive}
James~R Kubricht, Keith~J Holyoak, and Hongjing Lu,
\newblock ``Intuitive physics: Current research and controversies,''
\newblock {\em Trends in cognitive sciences}, vol. 21, no. 10, pp. 749--759,
  2017.

\bibitem{duchaine2009safe}
Vincent Duchaine and Cl{\'e}ment Gosselin,
\newblock ``Safe, stable and intuitive control for physical human-robot
  interaction,''
\newblock in {\em 2009 IEEE International Conference on Robotics and
  Automation}. IEEE, 2009, pp. 3383--3388.

\bibitem{zheng2015scene}
Bo~Zheng, Yibiao Zhao, Joey Yu, Katsushi Ikeuchi, and Song-Chun Zhu,
\newblock ``Scene understanding by reasoning stability and safety,''
\newblock {\em International Journal of Computer Vision}, vol. 112, no. 2, pp.
  221--238, 2015.

\bibitem{li2017visual}
Wenbin Li, Ale{\v{s}} Leonardis, and Mario Fritz,
\newblock ``Visual stability prediction for robotic manipulation,''
\newblock in {\em 2017 IEEE International Conference on Robotics and Automation
  (ICRA)}. IEEE, 2017, pp. 2606--2613.

\bibitem{forsyth2011computer}
David Forsyth and Jean Ponce,
\newblock {\em Computer vision: A modern approach.},
\newblock Prentice hall, 2011.

\bibitem{bengio2021deep}
Yoshua Bengio, Yann Lecun, and Geoffrey Hinton,
\newblock ``Deep learning for ai,''
\newblock {\em Communications of the ACM}, vol. 64, no. 7, pp. 58--65, 2021.

\bibitem{duan2021survey}
Jiafei Duan, Samson Yu, Hui~Li Tan, Hongyuan Zhu, and Cheston Tan,
\newblock ``A survey of embodied ai: From simulators to research tasks,''
\newblock {\em arXiv preprint arXiv:2103.04918}, 2021.

\bibitem{battaglia2016interaction}
Peter~W Battaglia, Razvan Pascanu, Matthew Lai, Danilo Rezende, and Koray
  Kavukcuoglu,
\newblock ``Interaction networks for learning about objects, relations and
  physics,''
\newblock {\em arXiv preprint arXiv:1612.00222}, 2016.

\bibitem{wu2016physics}
Jiajun Wu, Joseph~J Lim, Hongyi Zhang, Joshua~B Tenenbaum, and William~T
  Freeman,
\newblock ``Physics 101: Learning physical object properties from unlabeled
  videos.,''
\newblock in {\em BMVC}, 2016, vol.~2, p.~7.

\bibitem{lerer2016learning}
Adam Lerer, Sam Gross, and Rob Fergus,
\newblock ``Learning physical intuition of block towers by example,''
\newblock in {\em International conference on machine learning}. PMLR, 2016,
  pp. 430--438.

\bibitem{ye2018interpretable}
Tian Ye, Xiaolong Wang, James Davidson, and Abhinav Gupta,
\newblock ``Interpretable intuitive physics model,''
\newblock in {\em Proceedings of the European Conference on Computer Vision
  (ECCV)}, 2018, pp. 87--102.

\bibitem{groth2018shapestacks}
Oliver Groth, Fabian~B Fuchs, Ingmar Posner, and Andrea Vedaldi,
\newblock ``Shapestacks: Learning vision-based physical intuition for
  generalised object stacking,''
\newblock in {\em Proceedings of the European Conference on Computer Vision
  (ECCV)}, 2018, pp. 702--717.

\bibitem{leguen20phydnet}
Vincent Le~Guen and Nicolas Thome,
\newblock ``Disentangling physical dynamics from unknown factors for
  unsupervised video prediction,''
\newblock in {\em Computer Vision and Pattern Recognition (CVPR)}. 2020.

\bibitem{battaglia2013simulation}
Peter~W Battaglia, Jessica~B Hamrick, and Joshua~B Tenenbaum,
\newblock ``Simulation as an engine of physical scene understanding,''
\newblock {\em Proceedings of the National Academy of Sciences}, vol. 110, no.
  45, pp. 18327--18332, 2013.

\bibitem{fischer2016functional}
Jason Fischer, John~G Mikhael, Joshua~B Tenenbaum, and Nancy Kanwisher,
\newblock ``Functional neuroanatomy of intuitive physical inference,''
\newblock {\em Proceedings of the national academy of sciences}, vol. 113, no.
  34, pp. E5072--E5081, 2016.

\bibitem{ullman2017mind}
Tomer~D Ullman, Elizabeth Spelke, Peter Battaglia, and Joshua~B Tenenbaum,
\newblock ``Mind games: Game engines as an architecture for intuitive
  physics,''
\newblock {\em Trends in cognitive sciences}, vol. 21, no. 9, pp. 649--665,
  2017.

\bibitem{fleming2014visual}
Roland~W Fleming,
\newblock ``Visual perception of materials and their properties,''
\newblock {\em Vision research}, vol. 94, pp. 62--75, 2014.

\bibitem{rossi2018speed}
Federica Rossi, Elisa Montanaro, and Claudio de’Sperati,
\newblock ``Speed biases with real-life video clips,''
\newblock {\em Frontiers in integrative neuroscience}, vol. 12, pp. 11, 2018.

\bibitem{mitko2021striking}
Alex Mitko and Jason Fischer,
\newblock ``A striking take on mass inferences from collisions,''
\newblock {\em Journal of Vision}, vol. 21, no. 9, pp. 2812--2812, 2021.

\bibitem{mitko2020all}
Alex Mitko and Jason Fischer,
\newblock ``When it all falls down: the relationship between intuitive physics
  and spatial cognition,''
\newblock {\em Cognitive research: principles and implications}, vol. 5, pp.
  1--13, 2020.

\bibitem{gerstenberg2017intuitive}
Tobias Gerstenberg and Joshua~B Tenenbaum,
\newblock ``Intuitive theories,''
\newblock {\em Oxford handbook of causal reasoning}, pp. 515--548, 2017.

\bibitem{Hamrick2015ThinkAT}
Jessica~B. Hamrick, Kevin~A. Smith, Thomas~L. Griffiths, and Edward Vul,
\newblock ``Think again? the amount of mental simulation tracks uncertainty in
  the outcome,''
\newblock {\em Cognitive Science}, 2015.

\bibitem{Bramley2018IntuitiveEI}
Neil~R. Bramley, Tobias Gerstenberg, Joshua~B. Tenenbaum, and Todd~M. Gureckis,
\newblock ``Intuitive experimentation in the physical world,''
\newblock {\em Cognitive Psychology}, vol. 105, pp. 9--38, 2018.

\bibitem{LUDWINPEERY2021101396}
Ethan Ludwin-Peery, Neil~R. Bramley, Ernest Davis, and Todd~M. Gureckis,
\newblock ``Limits on simulation approaches in intuitive physics,''
\newblock {\em Cognitive Psychology}, vol. 127, pp. 101396, 2021.

\bibitem{smith2013sources}
Kevin~A Smith and Edward Vul,
\newblock ``Sources of uncertainty in intuitive physics,''
\newblock {\em Topics in cognitive science}, vol. 5, no. 1, pp. 185--199, 2013.

\bibitem{firestone2016seeing}
Chaz Firestone and Brian Scholl,
\newblock ``Seeing stability: Intuitive physics automatically guides selective
  attention,''
\newblock {\em Journal of Vision}, vol. 16, no. 12, pp. 689--689, 2016.

\bibitem{weissenborn2019scaling}
Dirk Weissenborn, Oscar T{\"a}ckstr{\"o}m, and Jakob Uszkoreit,
\newblock ``Scaling autoregressive video models,''
\newblock {\em arXiv preprint arXiv:1906.02634}, 2019.

\bibitem{yan2021videogpt}
Wilson Yan, Yunzhi Zhang, Pieter Abbeel, and Aravind Srinivas,
\newblock ``Videogpt: Video generation using vq-vae and transformers,''
\newblock {\em arXiv preprint arXiv:2104.10157}, 2021.

\bibitem{xingjian2015convolutional}
SHI Xingjian, Zhourong Chen, Hao Wang, Dit-Yan Yeung, Wai-Kin Wong, and
  Wang-chun Woo,
\newblock ``Convolutional lstm network: A machine learning approach for
  precipitation nowcasting,''
\newblock in {\em Advances in neural information processing systems}, 2015, pp.
  802--810.

\bibitem{guen2020disentangling}
Vincent~Le Guen and Nicolas Thome,
\newblock ``Disentangling physical dynamics from unknown factors for
  unsupervised video prediction,''
\newblock in {\em Proceedings of the IEEE/CVF Conference on Computer Vision and
  Pattern Recognition}, 2020, pp. 11474--11484.

\bibitem{bear2021physion}
Daniel~M Bear, Elias Wang, Damian Mrowca, Felix~J Binder, Hsiau-Yu~Fish Tung,
  RT~Pramod, Cameron Holdaway, Sirui Tao, Kevin Smith, Li~Fei-Fei, et~al.,
\newblock ``Physion: Evaluating physical prediction from vision in humans and
  machines,''
\newblock {\em arXiv preprint arXiv:2106.08261}, 2021.

\bibitem{duan2021space}
Jiafei Duan, Samson Yu, and Cheston Tan,
\newblock ``Space: A simulator for physical interactions and causal learning in
  3d environments,''
\newblock in {\em Proceedings of the IEEE/CVF International Conference on
  Computer Vision}, 2021, pp. 2058--2063.

\bibitem{dasgupta2021avoe}
Arijit Dasgupta, Jiafei Duan, Marcelo~H Ang~Jr, and Cheston Tan,
\newblock ``Avoe: A synthetic 3d dataset on understanding violation of
  expectation for artificial cognition,''
\newblock {\em arXiv preprint arXiv:2110.05836}, 2021.

\bibitem{brubaker2009estimating}
Marcus~A Brubaker, Leonid Sigal, and David~J Fleet,
\newblock ``Estimating contact dynamics,''
\newblock in {\em 2009 IEEE 12th International Conference on Computer Vision}.
  IEEE, 2009, pp. 2389--2396.

\bibitem{fragkiadaki2015learning}
Katerina Fragkiadaki, Pulkit Agrawal, Sergey Levine, and Jitendra Malik,
\newblock ``Learning visual predictive models of physics for playing
  billiards,''
\newblock {\em arXiv preprint arXiv:1511.07404}, 2015.

\bibitem{li2016fall}
Wenbin Li, Seyedmajid Azimi, Ale{\v{s}} Leonardis, and Mario Fritz,
\newblock ``To fall or not to fall: A visual approach to physical stability
  prediction,''
\newblock {\em arXiv preprint arXiv:1604.00066}, 2016.

\bibitem{finn2016unsupervised}
Chelsea Finn, Ian Goodfellow, and Sergey Levine,
\newblock ``Unsupervised learning for physical interaction through video
  prediction,''
\newblock {\em Advances in neural information processing systems}, vol. 29, pp.
  64--72, 2016.

\bibitem{baradel2019cophy}
Fabien Baradel, Natalia Neverova, Julien Mille, Greg Mori, and Christian Wolf,
\newblock ``Cophy: Counterfactual learning of physical dynamics,''
\newblock {\em arXiv preprint arXiv:1909.12000}, 2019.

\bibitem{He_2016_CVPR}
Kaiming He, Xiangyu Zhang, Shaoqing Ren, and Jian Sun,
\newblock ``Deep residual learning for image recognition,''
\newblock in {\em Proceedings of the IEEE Conference on Computer Vision and
  Pattern Recognition (CVPR)}, June 2016.

\bibitem{hara2018can}
Kensho Hara, Hirokatsu Kataoka, and Yutaka Satoh,
\newblock ``Can spatiotemporal 3d cnns retrace the history of 2d cnns and
  imagenet?,''
\newblock in {\em Proceedings of the IEEE conference on Computer Vision and
  Pattern Recognition}, 2018, pp. 6546--6555.

\bibitem{kataoka2020would}
Hirokatsu Kataoka, Tenga Wakamiya, Kensho Hara, and Yutaka Satoh,
\newblock ``Would mega-scale datasets further enhance spatiotemporal 3d
  cnns?,''
\newblock {\em arXiv preprint arXiv:2004.04968}, 2020.

\bibitem{devlin2018bert}
Jacob Devlin, Ming-Wei Chang, Kenton Lee, and Kristina Toutanova,
\newblock ``Bert: Pre-training of deep bidirectional transformers for language
  understanding,''
\newblock {\em arXiv preprint arXiv:1810.04805}, 2018.

\bibitem{subramanian2021spanpredict}
Vivek Subramanian, Matthew Engelhard, Sam Berchuck, Liqun Chen, Ricardo Henao,
  and Lawrence Carin,
\newblock ``Spanpredict: Extraction of predictive document spans with neural
  attention,''
\newblock in {\em Proceedings of the 2021 Conference of the North American
  Chapter of the Association for Computational Linguistics: Human Language
  Technologies}, 2021, pp. 5234--5258.

\bibitem{s20236941}
Xiang Yan, Syed~Zulqarnain Gilani, Mingtao Feng, Liang Zhang, Hanlin Qin, and
  Ajmal Mian,
\newblock ``Self-supervised learning to detect key frames in videos,''
\newblock {\em Sensors}, vol. 20, no. 23, 2020.

\bibitem{zhang2019spatio}
Ling Zhang, Le~Lu, Xiaosong Wang, Robert~M Zhu, Mohammadhadi Bagheri, Ronald~M
  Summers, and Jianhua Yao,
\newblock ``Spatio-temporal convolutional lstms for tumor growth prediction by
  learning 4d longitudinal patient data,''
\newblock {\em IEEE transactions on medical imaging}, vol. 39, no. 4, pp.
  1114--1126, 2019.

\bibitem{chai2021cms}
Zenghao Chai, Chun Yuan, Zhihui Lin, and Yunpeng Bai,
\newblock ``Cms-lstm: Context-embedding and multi-scale
  spatiotemporal-expression lstm for video prediction,''
\newblock {\em arXiv preprint arXiv:2102.03586}, 2021.

\bibitem{radford2015unsupervised}
Alec Radford, Luke Metz, and Soumith Chintala,
\newblock ``Unsupervised representation learning with deep convolutional
  generative adversarial networks,''
\newblock {\em arXiv preprint arXiv:1511.06434}, 2015.

\bibitem{6795228}
Ronald~J. Williams and David Zipser,
\newblock ``A learning algorithm for continually running fully recurrent neural
  networks,''
\newblock {\em Neural Computation}, vol. 1, no. 2, pp. 270--280, 1989.

\bibitem{wu2016google}
Yonghui Wu, Mike Schuster, Zhifeng Chen, Quoc~V Le, Mohammad Norouzi, Wolfgang
  Macherey, Maxim Krikun, Yuan Cao, Qin Gao, Klaus Macherey, et~al.,
\newblock ``Google's neural machine translation system: Bridging the gap
  between human and machine translation,''
\newblock {\em arXiv preprint arXiv:1609.08144}, 2016.

\bibitem{61115}
J.~Lin,
\newblock ``Divergence measures based on the shannon entropy,''
\newblock {\em IEEE Transactions on Information Theory}, vol. 37, no. 1, pp.
  145--151, 1991.

\bibitem{NEURIPS2019_9015}
Adam Paszke, Sam Gross, Francisco Massa, Adam Lerer, James Bradbury, Gregory
  Chanan, Trevor Killeen, Zeming Lin, Natalia Gimelshein, Luca Antiga, Alban
  Desmaison, Andreas Kopf, Edward Yang, Zachary DeVito, Martin Raison, Alykhan
  Tejani, Sasank Chilamkurthy, Benoit Steiner, Lu~Fang, Junjie Bai, and Soumith
  Chintala,
\newblock ``Pytorch: An imperative style, high-performance deep learning
  library,''
\newblock in {\em Advances in Neural Information Processing Systems 32},
  H.~Wallach, H.~Larochelle, A.~Beygelzimer, F.~d\textquotesingle
  Alch\'{e}-Buc, E.~Fox, and R.~Garnett, Eds., pp. 8024--8035. Curran
  Associates, Inc., 2019.

\bibitem{wolf2019huggingface}
Thomas Wolf, Lysandre Debut, Victor Sanh, Julien Chaumond, Clement Delangue,
  Anthony Moi, Pierric Cistac, Tim Rault, R{\'e}mi Louf, Morgan Funtowicz,
  et~al.,
\newblock ``Huggingface's transformers: State-of-the-art natural language
  processing,''
\newblock {\em arXiv preprint arXiv:1910.03771}, 2019.

\bibitem{kingma2014adam}
Diederik~P Kingma and Jimmy Ba,
\newblock ``Adam: A method for stochastic optimization,''
\newblock {\em arXiv preprint arXiv:1412.6980}, 2014.

\bibitem{geirhos2020shortcut}
Robert Geirhos, J{\"o}rn-Henrik Jacobsen, Claudio Michaelis, Richard Zemel,
  Wieland Brendel, Matthias Bethge, and Felix~A Wichmann,
\newblock ``Shortcut learning in deep neural networks,''
\newblock {\em Nature Machine Intelligence}, vol. 2, no. 11, pp. 665--673,
  2020.

\bibitem{standley2020tasks}
Trevor Standley, Amir Zamir, Dawn Chen, Leonidas Guibas, Jitendra Malik, and
  Silvio Savarese,
\newblock ``Which tasks should be learned together in multi-task learning?,''
\newblock in {\em International Conference on Machine Learning}. PMLR, 2020,
  pp. 9120--9132.

\end{thebibliography}

\newpage
\setcounter{section}{0}
\setcounter{table}{0}
\section*{Supplementary Material}
\section{Data Analysis}
To increase scale and complexity, we introduce the SPACE+ dataset, an improved and expanded version of the SPACE dataset \cite{duan2021space} with a larger dataset and additional unseen object classes respectively. Including data from the original SPACE dataset, we collect over 57,057 videos with over 8 million frames in total for seen object classes. We also collect an additional 11,411 videos for unseen object classes for a total of over 1.7 million frames.

The SPACE dataset comprises three novel video datasets that are synthesized based on three fundamental physical interactions, \emph{stability}, \emph{contact}, and \emph{containment}, in a 3D environment. Each interaction scenario is synthesized using the SPACE simulator, which is developed using Blender, an open-source 3D computer graphics tool with a Python API. The SPACE simulator generates the scenarios for the various physical interactions and their metadata for determining the outcomes of these physical interactions for individual objects. In total, there are 15,000 videos lasting 3 seconds with a 50 frames per second frame rate. Each video also comes with other data such as the segmentation map, optical flow map, depth map and surface normal vector map of each frame.
\setcounter{figure}{0}

\begin{figure}
    \centering
    \includegraphics[width=\linewidth]{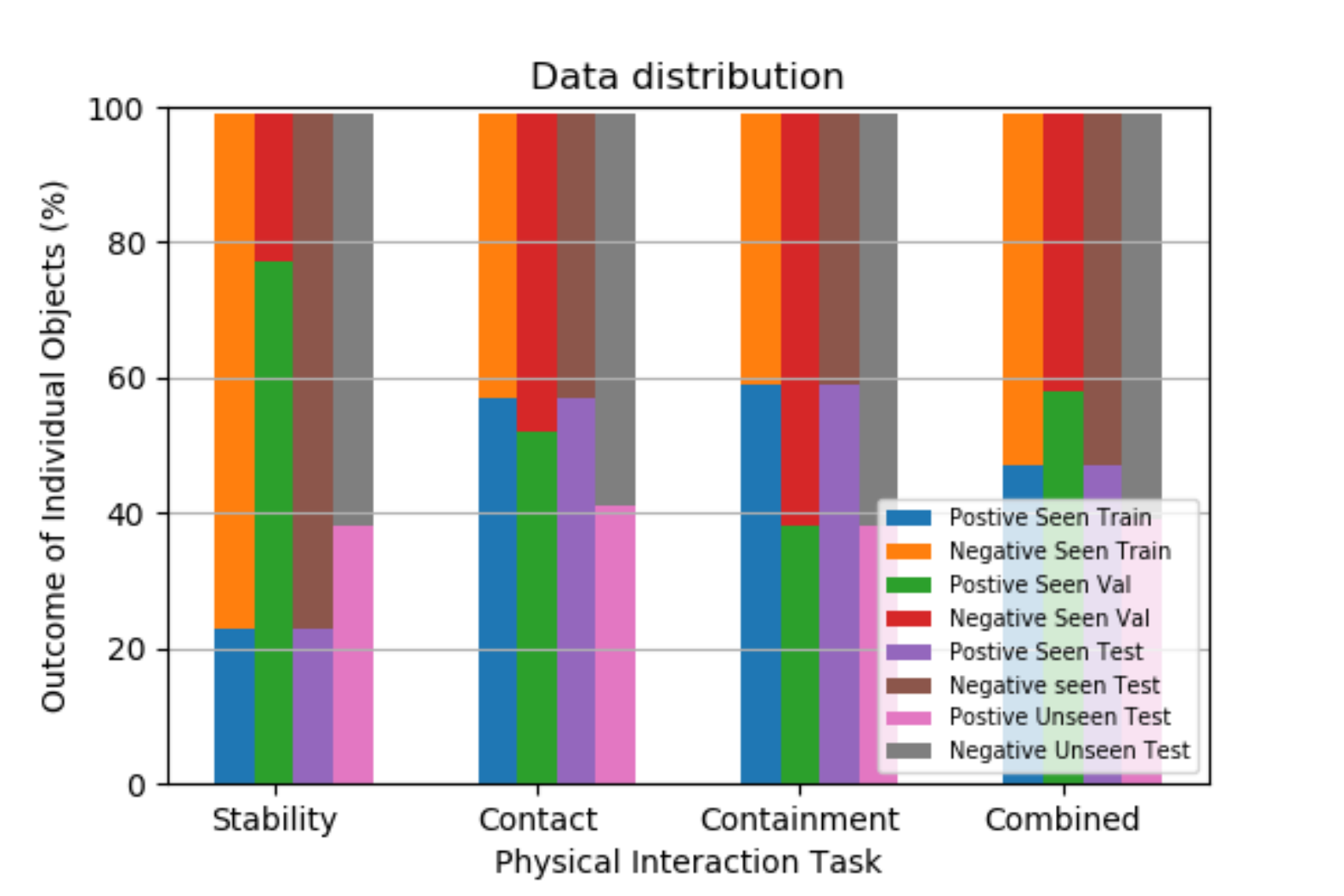} 
    \caption{Data distribution of the SPACE+ dataset used for training and testing in terms of physical interaction outcome.}
    \label{datadistribution}
\end{figure}

To train and test PIP, we utilize only 1,000 scenes from the SPACE+ dataset for each fundamental physical interaction task. We split the scenes for each task such that 60\% of the scenes (i.e. 600 scenes) are used for the training set and 20\% used for the validation and test sets each. For the combined task, each of the splits has equal numbers of each of the three fundamental tasks to ensure that the dataset is balanced. The data distribution of the 1,000 scenarios for each task in terms of physical interaction outcome is shown in Figure \ref{datadistribution}.

\begin{table*}[t]
\begin{center}
\resizebox{\linewidth}{!}{%
    \begin{tabular}{l|cccc}
    \hline
     & \multicolumn{4}{c}{Seen Objects (\%)} \\
     & \multicolumn{4}{c}{[Stability, Contact, Containment, Combined]} \\
    \hline
    Seed & {Baseline} & {PhyDNet} & {PIP w/o SS} & {PIP} \\
    \hline\hline
    1  & [92.35, 57.56, 78.87, 60.83] & [92.36, 68.77, 77.66, 58.85] & [91.70, 53.20, 84.30, 62.20] & [92.16, 89.65, 85.51, 79.13] \\
    2 & [92.35, 79.29, 82.87, 59.82] & [92.36, 53.62, 82.89, 65.94] & [92.30, 60.04, 80.08, 59.80] & [92.37, 87.79, 87.79, 76.76]  \\
    3 & [92.37, 68.12, 78.47, 60.24] & [92.36, 66.25, 76.25, 60.82] &  [92.30, 66.80, 84.50, 58.40] & [92.37, 86.96, 84.31, 80.71]\\
    4 &[92.35, 57.97, 68.41, 56.49] & [92.36, 57.97, 77.66, 59.00] & [92.30, 64.60, 80.28, 57.80] & [92.37, 87.37, 86.12, 78.74]  \\
    5 & [92.35, 65.21, 84.90, 66.53] & [92.36, 59.55, 82.89, 65.55] & [92.30, 64.60, 72.80, 55.30]& [92.37, 85.71, 88.53, 73.23]\\
    \hline
     \hline
     & \multicolumn{4}{c}{Unseen Objects (\%)} \\
     & \multicolumn{4}{c}{[Stability, Contact, Containment, Combined]} \\
    \hline
      Seed & {Baseline} & {PhyDNet} & {PIP w/o SS} & {PIP}\\
    \hline\hline
     1  & [65.38, 61.55, 59.24, 54.09] & [59.23, 42.71, 58.79, 59.00] & [64.60, 41.17, 64.10, 59.30] & [65.65, 63.81, 53.54, 68.98] \\
    2 & [69.23, 61.04, 56.00, 59.30] & [64.10, 66.58, 61.24, 49.37] & [68.90, 48.89, 54.12, 53.30] & [65.89, 61.80, 61.80, 57.57] \\
    3 & [65.12, 60.55, 56.34, 56.57] & [64.10, 66.08, 51.67, 64.51] &  [67.17, 73.86, 65.90, 49.80] &  [66.67, 58.04, 58.13, 66.50]\\
    4 &[56.39, 41.70, 54.12, 56.32] & [65.10, 41.70, 59.46, 52.60] & [69.70, 48.40, 57.24, 53.59]  & [67.44, 52.01, 52.56, 61.76]\\
    5 &[59.74, 45.47, 51.67, 48.63] & [66.40, 60.00, 64.36, 52.60] & [65.00, 48.40, 60.00, 57.00] & [66.41, 45.97, 53.90, 56.32]\\
    \hline
    
    \end{tabular}
    }
\end{center}
\caption{Results for all five seeds used for both seen (\emph{top}) and unseen (\emph{bottom}) object scenarios in our four physical interaction outcome prediction tasks.}
\label{allresults}
\end{table*}

\section{Results and Analysis}

\subsection{Detailed Test Results}
We present detailed test results for the baseline, modified PhyDNet and ablation (PIP w/o SS) models and PIP for all five seeds used for both seen and unseen object scenarios in Table \ref{allresults}.

\subsection{Cross-task Evaluation}

\begin{figure}
    \centering
    \includegraphics[width=\linewidth]{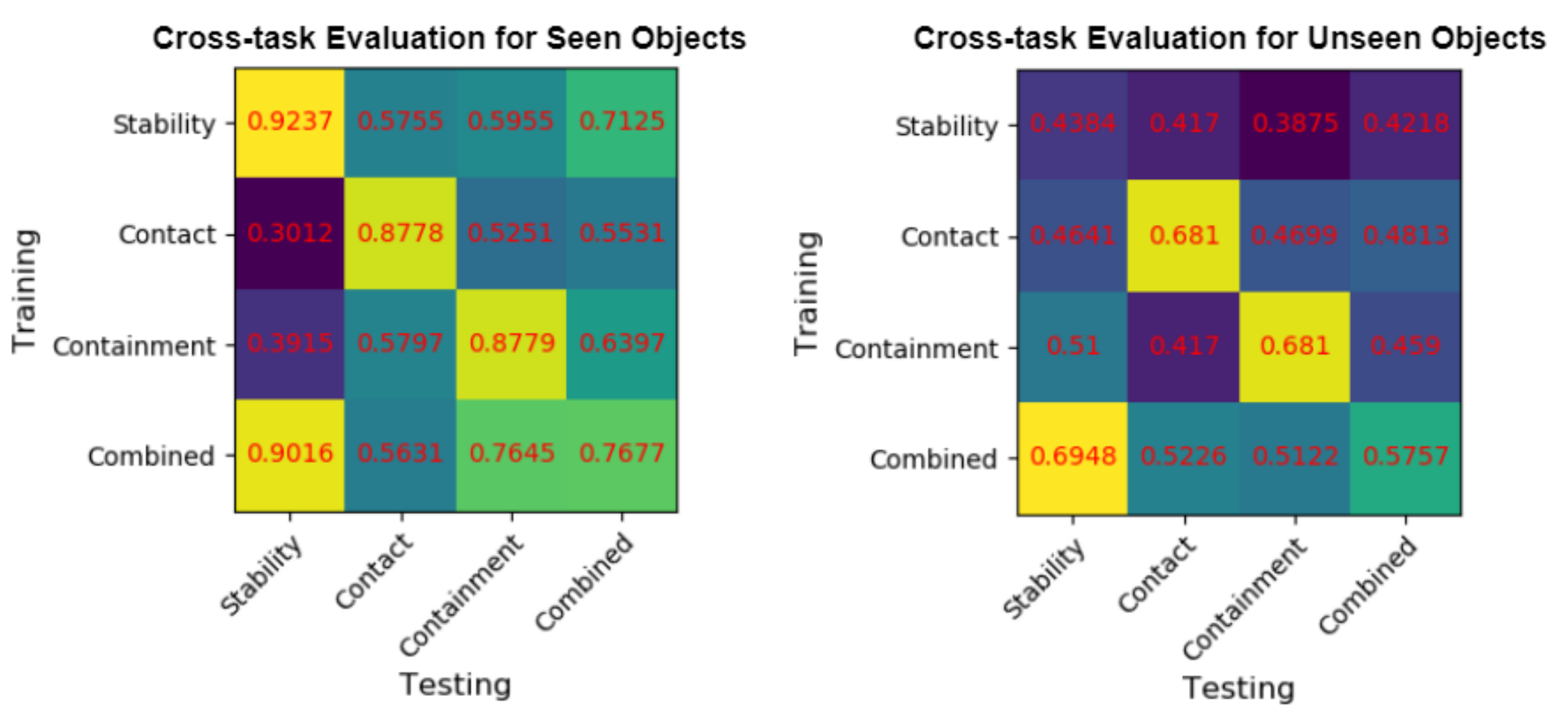} 
    \caption{Cross-task results for both seen (\emph{left}) and unseen (\emph{right}) objects for one seed run with PIP.}
    \label{cross}
\end{figure}

We conduct cross-task evaluation for PIP with one seed. The results are shown in Figure \ref{cross}. The results show that when PIP is trained on the combined task, it generally performs worse on each fundamental physical interaction task as compared to the cases where PIP is trained and tested on the same fundamental physical interaction task. This is a common problem in multi-task learning \cite{standley2020tasks}, and we aim to address this in future work.

\subsection{Span Selection Threshold}

We calculate the threshold for span selection as such:
\begin{gather*}
\mathrm{\textbf{p\_{threshold}}} = \mathrm{cumsum}([\frac{1}{N},\frac{1}{N},...,\frac{1}{N}]) \in \mathbb{R}^N \\
\mathrm{\textbf{q\_threshold}} = \mathrm{\textbf{p\_threshold}_{::-1}} \\
\mathrm{\textbf{r\_threshold}} = \frac{\mathrm{\textbf{p\_{threshold}}} \odot \mathrm{\textbf{q\_{threshold}}}}{\Sigma_t(\mathrm{\textbf{p\_{threshold}}}\odot \mathrm{\textbf{q\_{threshold}})_t}},
\end{gather*}
\begin{figure}
    \centering
    \includegraphics[width=0.8\linewidth]{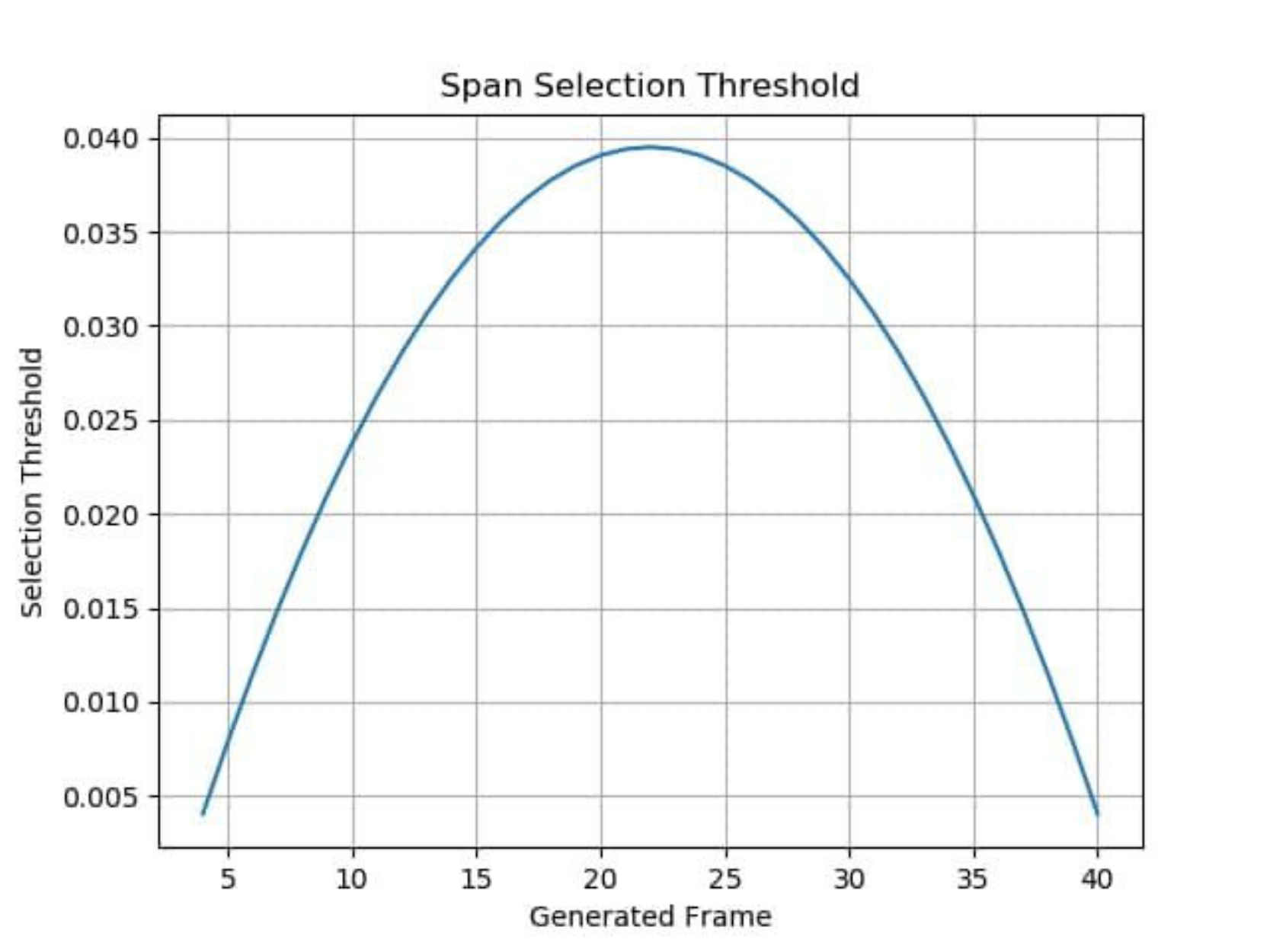} 
    \caption{Span selection thresholds for 37 generated frames.}
    \label{spanselectionthreshold}
\end{figure}

where $N$ is the generated frame sequence length. We show in Figure \ref{spanselectionthreshold} the span selection values for each frame for our generated frame sequence of length 37.

\subsection{PIP Test Visualizations}
\begin{figure*}[t]
    \centering
    \includegraphics[width=\linewidth]{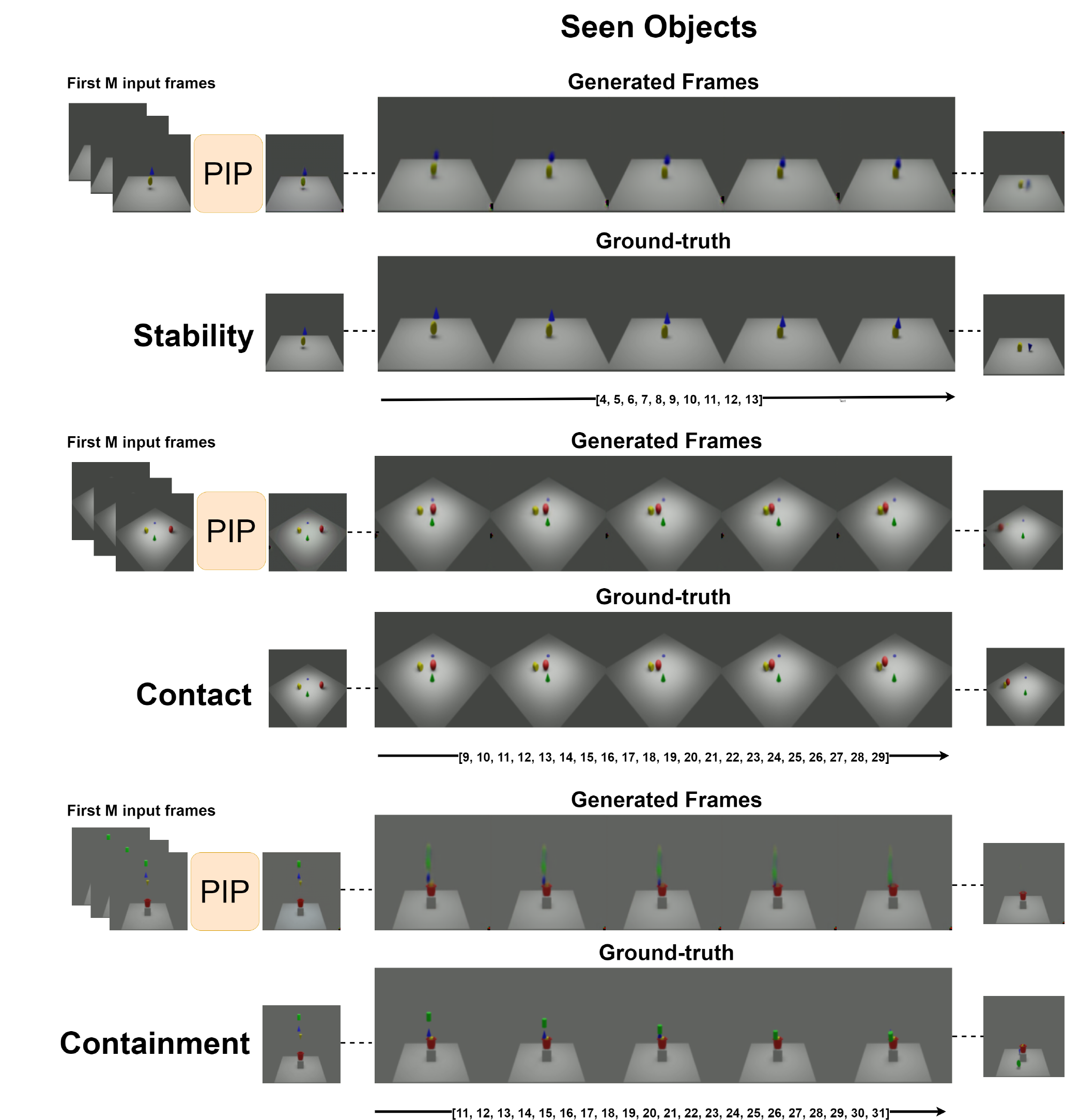} 
    \caption{Examples of generation and span selection with PIP for seen objects.}
    \label{spanseen}
\end{figure*}

\begin{figure*}[t]
    \centering
    \includegraphics[width=\linewidth]{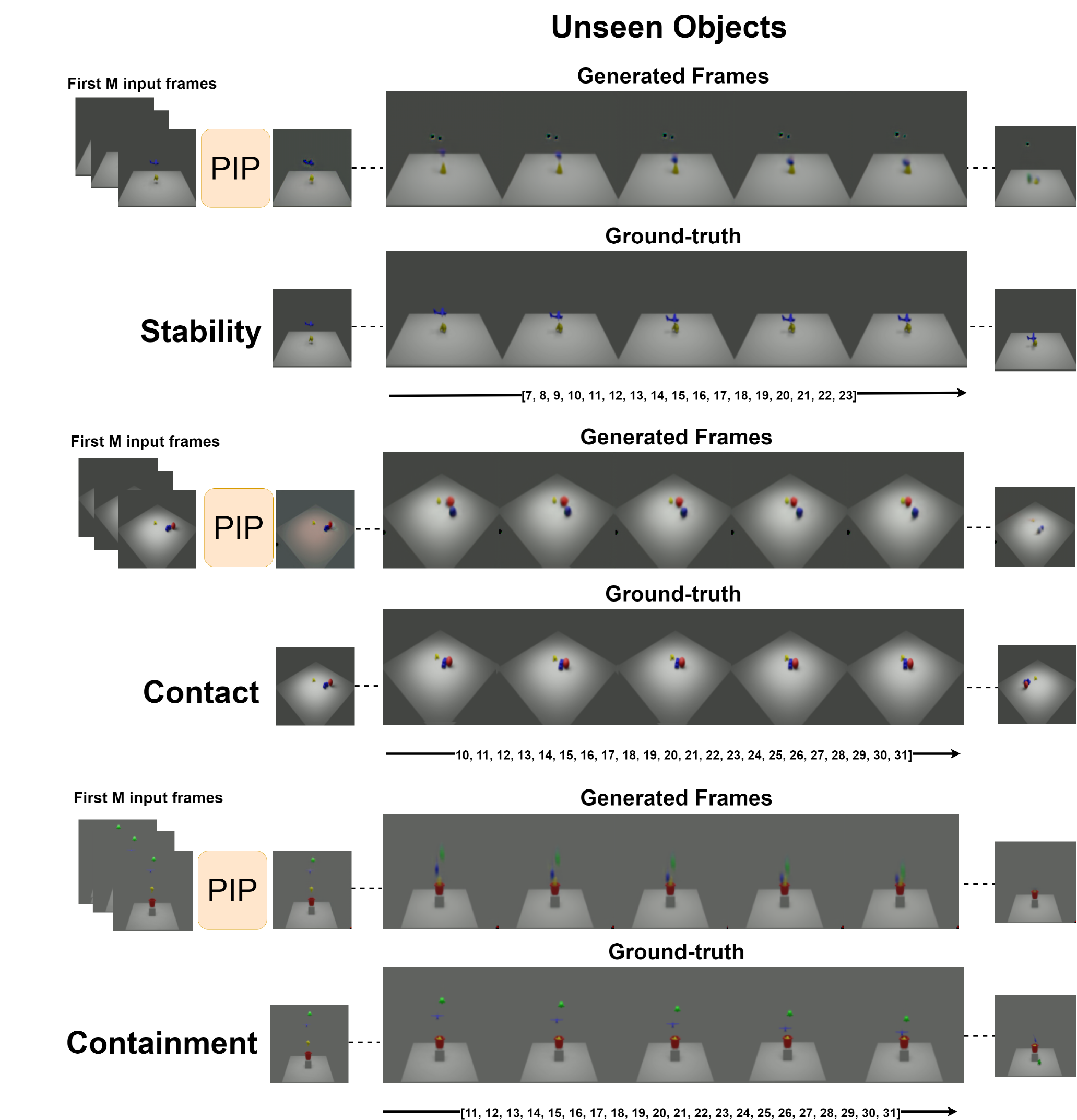} 
    \caption{Examples of generation and span selection with PIP for unseen objects.}
    \label{spanunseen}
\end{figure*}

We evaluate PIP on both seen and unseen object scenarios and extract several generation and span selection examples. The examples are shown in Figure \ref{spanseen} for seen objects and Figure \ref{spanunseen} for unseen objects. 

\begin{table}[t]
    \centering
    \resizebox{\linewidth}{!}{
    \begin{tabular}{ccc}
    \hline
    Name & URL & License \\
    \hline
    ConvLSTM \cite{xingjian2015convolutional} & \url{https://github.com/xibinyue/ConvLSTM-1} & GNU General Public License v3.0 \\
    3D ResNet \cite{kataoka2020would} & \url{https://github.com/kenshohara/3D-ResNets-PyTorch} & MIT License \\
    PhyDNet \cite{guen2020disentangling} &
    \url{https://github.com/vincent-leguen/PhyDNet} & GNU General Public License v3.0 \\
    SPACE \cite{duan2021space} & \url{https://github.com/jiafei1224/SPACE} & GNU General Public License v3.0 \\
    \hline
    \end{tabular}}
\caption{Open source code used.}
\label{opensource}
\end{table}

\section{Code and Software}
We use these software libraries and their versions:
\begin{itemize}
  \item matplotlib: 3.3.4
  \item natsort: 7.1.1
  \item numpy: 1.20.2
  \item opencv-python: 4.5.2.54
  \item piqa: 1.1.3
  \item pytorch: 1.8.0
  \item scikit-image: 0.18.1
  \item scipy: 1.6.1
  \item tqdm: 4.59.0
  \item transformers: 4.9.2
  \item yaml: 0.2.5
\end{itemize}

The open source code we use can be found in Table \ref{opensource}.
\clearpage
\end{document}